\documentclass{article}
\usepackage[utf8]{inputenc}
\usepackage{geometry}
\usepackage{amsmath}
\usepackage{amssymb}
\usepackage{graphics}
\usepackage{graphicx}
 \usepackage{subfigure} 
\title{Predictive Modeling of Menstrual Cycle Length: A Time Series Forecasting Approach}
\author{Prof. Dr. Rosana Rego\\
Department of Engineering and Technology \\
Federal University of Semi-Arid Region\\
rosana.rego@ufersa.edu.br}
\date{}

\begin{document}

\maketitle

\noindent \textbf{Abstract.} A proper forecast of the menstrual cycle is meaningful for women's health, as it allows individuals to take preventive actions to minimize cycle-associated discomforts. In addition, precise prediction can be useful for planning important events in a woman's life, such as family planning. In this work, we explored the use of machine learning techniques to predict regular and irregular menstrual cycles. We implemented some time series forecasting algorithm approaches, such as AutoRegressive Integrated Moving Average, Huber Regression, Lasso Regression, Orthogonal Matching Pursuit, and Long Short-Term Memory Network. Moreover, we generated synthetic data to achieve our purposes. The results showed that it is possible to accurately predict the onset and duration of menstrual cycles using machine learning techniques.

\section{Introduction}

The menstrual cycle is the time from the first day of a woman's period to the day before her next period \cite{campbell2021menstrual}.  It differs from woman to woman, in addition, it is connected to physical and mental well-being \cite{tatsumi2020age}. The correct knowledge of cycle variations is crucial to understanding reproductive health and defining multidimensional well-being in women \cite{tatsumi2020age}. Some phases constitute the menstrual cycle, such as menstrual phase or period, follicular phase, ovulation phase, and luteal phase, sequentially \cite{trickey2004women}. The length of each phase can vary from woman to woman, and it can modify over time. In this way, a cycle is considered regular when it varies from 21 to 35 days, counting from the first day of one period to the first day of the next \cite{trickey2004women}. In a regular cycle, the period can endure between three and five days \cite{trickey2004women}. The luteal phase, usually, will be fourteen days long, and the follicular phase will also be fourteen days long \cite{trickey2004women}. However, not all woman will have this menstrual pattern. During the woman's life, the periods can alter, i. e., they may stay longer or get lighter \cite{bull2019real}. The periods will persist until menopause, which usually happens when women are in their late 50s  \cite{trickey2004women}. 
Moreover, the cycle can be irregular.  A cycle is considered irregular when most of previous cycles are outside regular intervals, i.e. there is no visible pattern in the variation \cite{rees2005abnormal}.

Irregular cycles are common and usually not a cause for worry. Factors that can contribute to irregular periods include: natural hormonal changes, hormonal birth control, stress, resistance exercise, and weight loss \cite{munro2012figo}. For instance, for a woman in the perimenopause irregularity of periods is extremely common. However, an irregular cycle may be a sign of either early ovarian failure or polyscystic ovary syndrome \cite{rees2005abnormal, harlow1995epidemiology}. Moreover, irregular cycles can sometimes drive pregnancy complicated, as the days a woman is ovulating can also be irregular  \cite{campbell2021menstrual}.
In this way, women who hold irregular cycles and desire to become pregnant have traditionally resorted to the arduous strategy of tracking urinary luteinizing hormone (LH), measuring basal body temperature (BBT), calendars and calculations to estimate the ovulatory window and thus maximizing  their chances of getting pregnant \cite{rostvik2022cash}. However, manually mapping and tracking fertility through BBT and LH tests can be arduous \cite{yu2022tracking}.

The knowledge of the menstrual cycle allow women understanding their own body and may detected a possible menstrual disorders \cite{nguyen2021detecting}. However, many women are troubled by an irregular or unpredictable menstrual cycle \cite{kwak2018irregular}. The patterns are so individual to each woman that it is hard for a human to create enough rules to capture them. In this way, machine learning (ML) techniques can be applied to predict a woman's cycle. ML and deep learning algorithms can detect each woman's unique and individual patterns and better predict the ovulation window and period. Computational intelligence can be helpful for disease diagnosis, decision support systems, and prognosis \cite{tiwari2022sposds, urteaga2021generative, pearson2021natural}. Because of this, it has become an important approach in biomedical discovery and clinical research.

ML algorithms, such as artificial neural networks (ANN) can deliver significantly better solutions than known methods to cycle predictions \cite{li2022predictive}. As described by \cite{shinde2018intelligent},  machine learning procedures enable the development of intelligence into a machine so that it can achieve better in the future using the learned experience. Moreover, machine learning appliances carry about smart change in the health industry, which includes pattern detection \cite{alhussein2018voice},  predictions system \cite{nithya2017predictive,sarwar2018prediction}, image recognition \cite{rahane2018lung}. In this way, in this paper, we highlighted the prediction system for the predictive modeling of healthcare, particularly menstrual cycle length from synthetic data.

One of the prior challenges in predictive modeling for the menstrual cycle pertains to addressing the uncertainty of the data at hand, as well as ensuring the adjustment of the model output \cite{urteaga2021generative, li2022predictive, denny2019hope}. In manuscript \cite{urteaga2021generative}, the authors explored uncertainty quantification when designing predictive models for healthcare, such as cycle length patterns. In the same way, \cite{li2022predictive} proposed a predictive model for predict the next menstrual cycle start date. Differently, in \cite{denny2019hope}, the authors applied some machine learning models to predict polycystic ovary syndrome. Therefore, these works have demonstrated the capability of probabilistic models and machine learning algorithms to detect patterns in women's cycles.

Motivated by the above discussion, our purpose is to accurately forecast the subsequent cycle based on previously tracked cycle and periods.  Thus, the main contributions of this work lie in the following: 
\begin{itemize}
    \item [i.]  Use machine learning algorithms to analyze data from previous menstrual cycles in order to identify patterns and predict the start of future cycles;
    \item[ii.] Model that can accurately predict the timing and length of the menstrual cycle using only historical menstrual cycle data.
\end{itemize}

The manuscript is divided into: The introduction, where we provided background information on the problem and its significance. The related work section summarizes the relevant research on the topic.
The problem formulation section describes the problem and some time series algorithms approaches. The data characteristics section describes the research design and data generation methods used in our study. The results section presents our findings, including statistical analyses and visual representations of the data. Finally, the conclusion summarizes the main findings and contributions of our study, and suggests directions for future research.

\section{Related work}

There is a need for the development of an appropriate model for one-step-ahead forecast periods.In the last few years, the usage of web systems and apps to track periods and predict the fertile window has become popular \cite{setton2016accuracy, deverashetti2022security}. For instance, in the manuscript \cite{de2021modelling}, the authors built a hybrid predictive model based on autoregressive–moving-average (ARMA) and linear mixed effect model (LMM) for forecasting menstrual cycle length in athletes. In paper 
\cite{fukaya2017forecasting}, the authors presented a state-space modeling approach for forecasting the onset of menstruation based on basal body temperature time series data. They used the state-space model to forecast the onset of menstruation for up to three months in advance. Differently, in \cite{chen2014mathematical}, Chen et al.  proposed a mathematical model of the human menstrual cycle. The model is based on a system of differential equations that describes the changes in hormone levels in the body over the course of the menstrual cycle.  The authors validated the model using data from clinical studies of the menstrual cycle, and found that it accurately predicted the timing and magnitude of hormone fluctuations during the cycle. 

In \cite{roblitz2013mathematical}, the authors presented a mathematical model for the administration of gonadotropin-releasing hormone (GnRH) analogues in the treatment of infertility. The authors developed the model to optimize the timing and dosage of GnRH analogue administration for the induction of ovulation. The model takes into account the feedback mechanisms that regulate hormone levels, as well as the effects of GnRH analogues on hormone production.

To enhance analysis, we compiled some manuscripts that modeling cycles in time series data. In Table \ref{tab}, we summarized the paper, model used by authors, and features employed.

\begin{table}[!htb]
\begin{tabular}{lll}
Paper & Model & Features \\
\hline
   \cite{de2021modelling}   &   ARMA, LMM & Cycle lengths\\

   \cite{li2021generative} & Poisson model & Cycle lengths \\
   
   \cite{urteaga2021generative}   & CNN, RNN, LSTM, Generalized Poisson   & Cycle lengths     \\
   \cite{pierson2018modeling} & Cyclic Hidden Markov Models & Cycle lengths\\
    \cite{kleinschmidt2019advantages}  &     Natural Cycles algorithm*  & Cycle lengths, BBT, LH     \\   
\end{tabular}

 {\raggedright \small{*Algorithm information are not provided.}\par}
 \caption{Relevant papers.}
    \label{tab}
\end{table}

The authors, in \cite{de2021modelling}, employed the ARMA, and LMM models, focusing on cycle lengths as the key feature of interest. \cite{li2021generative} used a Poisson model, with a focus on cycle lengths as well. In \cite{urteaga2021generative} a comparison between  Convolutional Neural Network (CNN), Recurrent Neural Network (RNN), Long Short-Term Memory (LSTM), and Generalized Poisson models are made.  \cite{kleinschmidt2019advantages} compared fertile window and cycle predicted by Natural Cycles algorithm over calendar-based methods. The Natural Cycle algorithm incorporated cycle lengths, BBT, and LH as key features for analysis. 

In contrast to previous works, in this study, we implemented multiple models to forecast the menstrual cycle duration. These models include AutoRegressive Integrated Moving Average (ARIMA), Huber Regression, which is a robust regression method that provides robustness against outliers, Least Absolute Shrinkage and Selection Operator (LASSO) Regression, Orthogonal Matching Pursuit (OMP), which efficiently identifies relevant predictors, and LSTM Network. 

\section{Problem formulation}

The problem formulation for predicting menstrual cycle based on only the cycle time series, involves developing a model that can accurately predict the timing and length of the menstrual cycle using only historical menstrual cycle data. The model is designed to handle both regular and irregular cycles, and may need to be personalized to the individual woman. 

The menstrual cycle time series can be represented as a sequence of observations indexed by time, where each observation corresponds to a menstrual cycle.  
Let us denote the menstrual cycle length time series by 

\begin{equation}
    X_1 = x_1(t),
\end{equation}
where $x_1(t)$ represents the length of the menstrual cycle at time $t$.
Similarly, let's denote the period length time series by 
\begin{equation}
    X_2 = x_2(t),
\end{equation}
where $x_2(t)$  represents the length of the bleeding time at time $t$.

We can use a sequence-to-sequence model, where the input sequence consists of past values of $X_1$  and $X_2$, and the output sequence consists of future values of $X_1$ and $X_2$. Therefore, let us denote the input sequence $X \in R^{N\times 2}$ as 

\begin{equation}
    X = \left[ \begin{array}{cc}
        x_1(t-L+1) & x_2(t-L+1)\\
        \vdots &  \vdots \\
        x_1(t) &  x_2(t) 
    \end{array} \right],
\end{equation}
where $L$  is the length of the input sequence. 

We can write the output sequence $Y R^{N\times 2}$ as 
\begin{equation}
    Y = \left[ \begin{array}{cc}
        x_1(t+1) & x_2(t+1)\\
        \vdots &  \vdots \\
        x_1(t+P) &  x_2(t+P) 
    \end{array} \right],
\end{equation}
where $P$ is the length of the output sequence.  Each element of the matrices $X$ and $Y$  corresponds to a time step and consists of the menstrual cycle length and period length at that time step.

\subsection{Time series algorithms approaches}

The model must be able to identify the key patterns and trends in the menstrual cycle time series data, such as the length of the menstrual cycle, and the duration of menstruation. Machine learning algorithms, such as recurrent neural networks or time-series forecasting models, such as ARIMA, can be used to analyze the menstrual cycle time series and predict future menstrual cycles based on past cycles.  The model can also be used to identify any irregularities or abnormalities in the menstrual cycle, which can aid in the diagnosis and treatment of reproductive health issues.

In the following sections, we discussed some models that have been utilized, including the LSTM network, ARIMA, Huber Regression, Lasso Regression, and OMP.

\subsubsection{LSTM model} 

Long Short-Term Memory (LSTM) is a type of recurrent neural network that is generally employed for sequential data processing, such as time series analysis, and natural language processing \cite{hochreiter1997long}. The fundamental idea behind LSTM is the usage of memory cells, which are capable of maintaining information over long sequences \cite{hochreiter1997long}.

We applied an LSTM network to predict the menstrual cycle time series.  The LSTM model takes the input sequence $X$  and outputs a sequence of predicted values 

\begin{equation}
   \hat{ Y }= \left[ \begin{array}{cc}
        x_1(t+1) & x_2(t+1)\\
        \vdots &  \vdots \\
        x_1(t+P) &  x_2(t+P) 
    \end{array} \right],
\end{equation}
with $\hat{ Y } \in R^{N\times 2}$. The LSTM model can be formulated as follows:

\begin{equation}
    \begin{array}{c}
       h(t) = LSTM(X(t), h(t-1)) \\
\hat{ Y }(t) = linear(h(t))
    \end{array}
\end{equation}
where $h(t)$ is the hidden state at time $t$, which is passed to the output layer to predict the values of $\hat{ Y }$, and linear is a linear transformation to map the hidden state to the output values.

The LSTM cell updates the hidden state based on the current input and the previous hidden state as follows:

\begin{equation}
  \begin{array}{c}
  i(t) = \varphi(W_i [h(t-1), X(t)] + b_i) \\
  f(t) = \varphi(W_f  [h(t-1), X(t)] + b_f) \\
o(t) = \varphi(W_o  [h(t-1), X(t)] + b_o) \\
\hat{c}(t) = tanh(W_c  [h(t-1), X(t)] + b_c) \\
c(t) = f(t)  c(t-1) + i(t)  \hat{c}(t) \\
h(t) = o(t)  tanh(c(t))
  \end{array}
\end{equation}
where $i(t)$, $f(t)$, $o(t)$ are the input, forget, and output gates, respectively, $W_i$, $W_f$, $W_o$, $W_c$ are weight matrices, $b_i$, $b_f$, $b_o$, $b_c$ are bias vectors, and $c(t)$ is the cell state at time t. The output layer is a linear layer that takes the hidden state $h(t)$ as input and outputs the predicted values $\hat{ Y }(t)$.

\subsubsection{ARIMA model}

ARIMA model is a time series forecasting model that incorporates autoregression and moving average components \cite{chatfield2000time}. We can expressed it as

\begin{equation}
   {\hat{ Y }'}_t = c + \sum(\phi_i   {\hat{ Y }'}_{t-i}) +  \sum(\theta_i  \epsilon_{t-i}) + \epsilon_t
\end{equation}
where $  {\hat{ Y }'}_t $  represents the differenced  observed menstrual cycle length at time $t$, $c$ is a constant term, $\phi$ represents the coefficient of the autoregression component at lag $i$. Note that, we used the past values of the differenced time series to predict future values. $\epsilon_t$  is the white noise error term at time t, and $\theta_i$ represents the coefficient of the moving average component at lag $i$.

ARIMA models can be used to predict the cycles patterns in a time series by incorporating seasonal differencing and seasonal autoregressive and moving average components. 

\subsubsection{Huber Regression}

Huber regression is a robust regression that uses the Huber loss function. Unlike the squared error loss, the Huber loss function is designed to be less influenced by outliers \cite{huber1992robust}. Its definition can be expressed as follows:

\begin{equation}
 L_{\delta} ({Y}, X\beta ) =    \left\{ \begin{array}{cr}
   \frac{1}{2} ({Y}- X\beta )^2 & \text{for } |{Y}- X\beta | \leq \delta \\
   \delta (| {Y}- X\beta | - \frac{1}{2} \delta) & \text{otherwise.}
    \end{array}\right.
\end{equation}
where $\beta$ is the regression coefficients, and $\delta$ is a tuning parameter, which determines the threshold beyond which the loss function transitions from squared to absolute loss, thereby controlling the model's sensitivity to outliers \cite{huber1992robust}. Therefore the problem consist in 

\begin{equation}
    min \sum_{i}{L_{\delta} ({Y}_i, X_i\beta )}.
\end{equation}
Hence, the goal is to encounter the regression coefficients $\beta$ that minimize the sum of the Huber losses between the predicted values  $X_i\beta$ and the actual values ${Y}_i$. 

\subsubsection{Lasso regression}

Unlike Huber regression, Lasso regression utilizes the L1 norm penalty (sum of absolute values) on the regression coefficient, i.e., Lasso regression promotes sparsity in the regression coefficients by adding the penalty \cite{tibshirani1996regression}. The minimization problem for Lasso regression can be stated as follows:

\begin{equation}
    min \frac{1}{2n}\sum_{i}{ ({Y}_i - X_i\alpha )^2} + \lambda\sum_{j}{|\alpha_j|}  .
\end{equation}
where $\lambda$ is the regularization parameter that controls the strength of the penalty term, and $\alpha$  is the vector of regression coefficients.

\subsubsection{Orthogonal Matching Pursuit}

OMP is an applied algorithm for recovering sparse high-dimensional vectors in linear regression models \cite{kallummil2018signal}. The objective of OMP in cycles time series forecasting is to determine the lagged variables, i.e., the predictors that are most informative in describing the current and future values of a cycle. Therefore, given the time series $X$, and a set of lagged variables $\hat{Y}$ \cite{cai2011orthogonal}:

\begin{itemize}
    \item [1.] Initialize the set of selected predictors $S = \emptyset$ and the residual $r = X$. Let the iteration counter $i=1.$
    \item [2.] Find $\hat{Y}_i$ that solves 
    $ max |\hat{Y}^T r_{i-1}|$
and add  $\hat{Y}_i$ to the set $S$.
\item [3.] Let $P_i= \hat{Y}(\hat{Y}^T\hat{Y})^{-1} \hat{Y}^T$ denote the projection into the linear space spanned by the elements of $\hat{Y}$. Update $r_i=(I-P_i)X.$ 
\item [4.] If the stopping condition is achieved, stop the algorithm. Otherwise, set $i=i+2$ and return to Step 2. 

\end{itemize}

OMP algorithm was adjusted to combine lagged variables that capture the temporal dependencies in the data. By iteratively selecting lagged variables based on their correlations with the temporal residuals, OMP gradually determines the most informative predictors for prediction purposes in the cycle time series.

\section{Data Characteristics }

We generated synthetic data that mimic the statistical properties, such as the distribution, variance of a real cycle. This is useful for exploring different scenarios and testing the robustness of the machine learning models to different data conditions. 
The data is generated from the following model:

\begin{equation} \label{eq:1}
\begin{split} \xi
 = \xi_{mean} + \xi_{std} \cdot \mu_\xi \\  \varrho = \varrho_{mean} + \varrho_{std} \cdot \mu_\varrho
\end{split} \end{equation}
where $\varrho$ represents the period length, $\xi$ is the cycle duration in days, and $\mu$ denotes the uncertain.  

According to the model, the cycle length is generated based on the mean and the standard deviation of the cycle length. The period length is generated based on the mean and the standard deviation of the period length. 
For analysis of the possible patterns present in regular and irregular cycles, we explored three different cases: 

\begin{itemize} \item  \textbf{Case 1:} The period starts on the first day of the cycle, and the variation of the period length is small, i.e., $\varrho_{std} \ll \varrho_{mean}$. Therefore, it is a regular cycle.

\item \textbf{Case 2:} The period starts on the last day of the cycle, and the variation of the cycle is between 28 and 35 days, and a period variation between 5 and 6 days. Therefore, it is a regular cycle.

\item \textbf{Case 3:} The period starts in the middle of the cycle, and the variation of the cycle is between 28 and 49 days and a period variation between 4 and 8 days. Therefore, it is an irregular cycle. 
\end{itemize}

\begin{figure}[!htb]
    \centering
 \subfigure{ \includegraphics[scale=0.3]{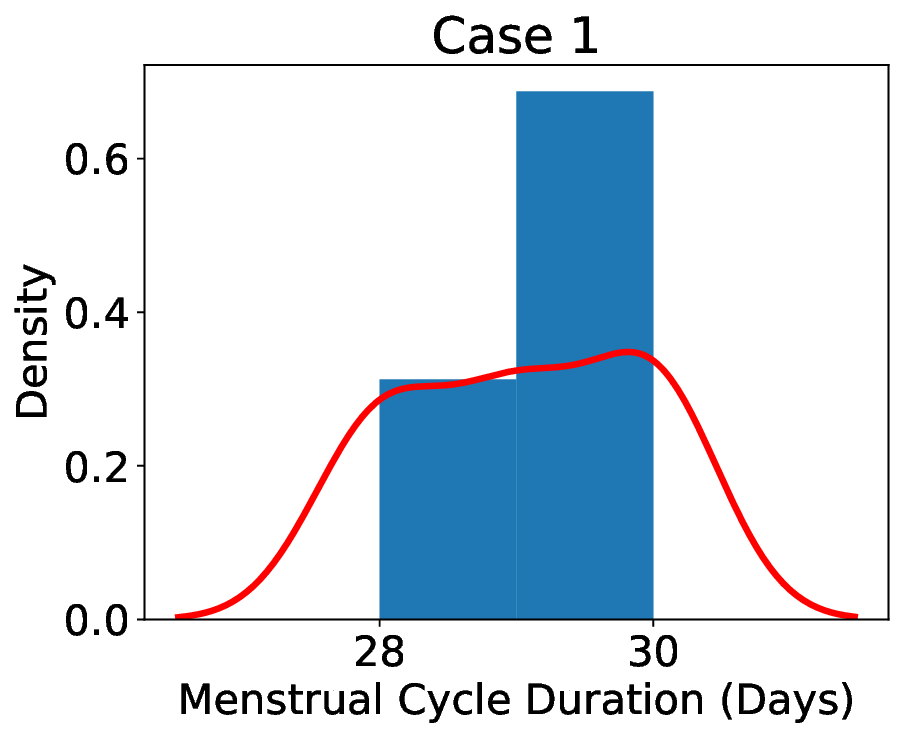}}
 \subfigure{ \includegraphics[scale=0.3]{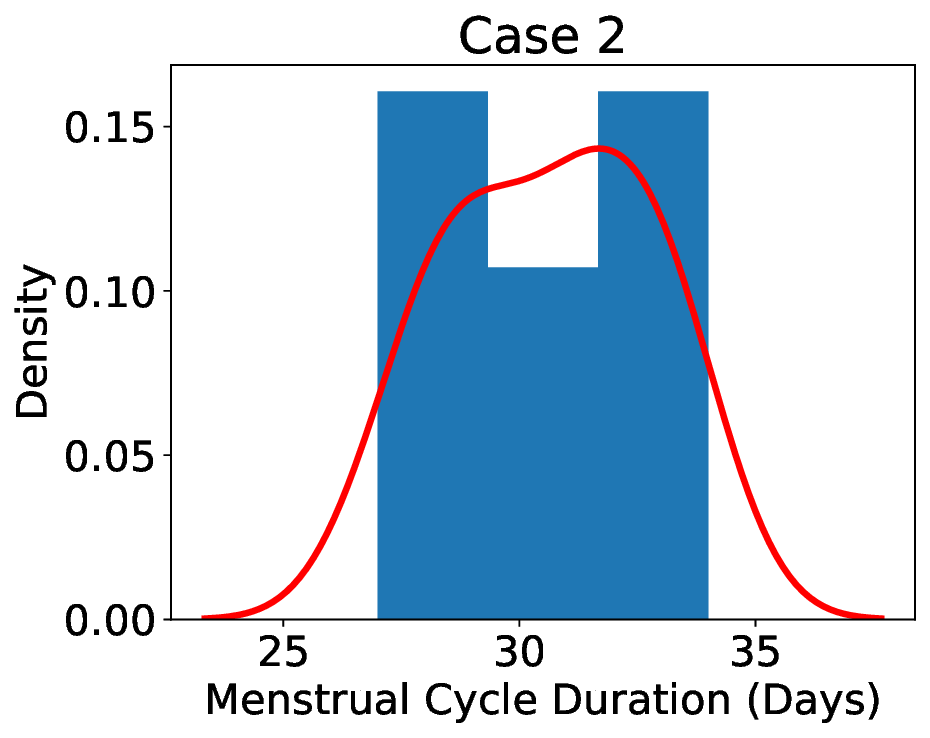}}
 \subfigure{ \includegraphics[scale=0.3]{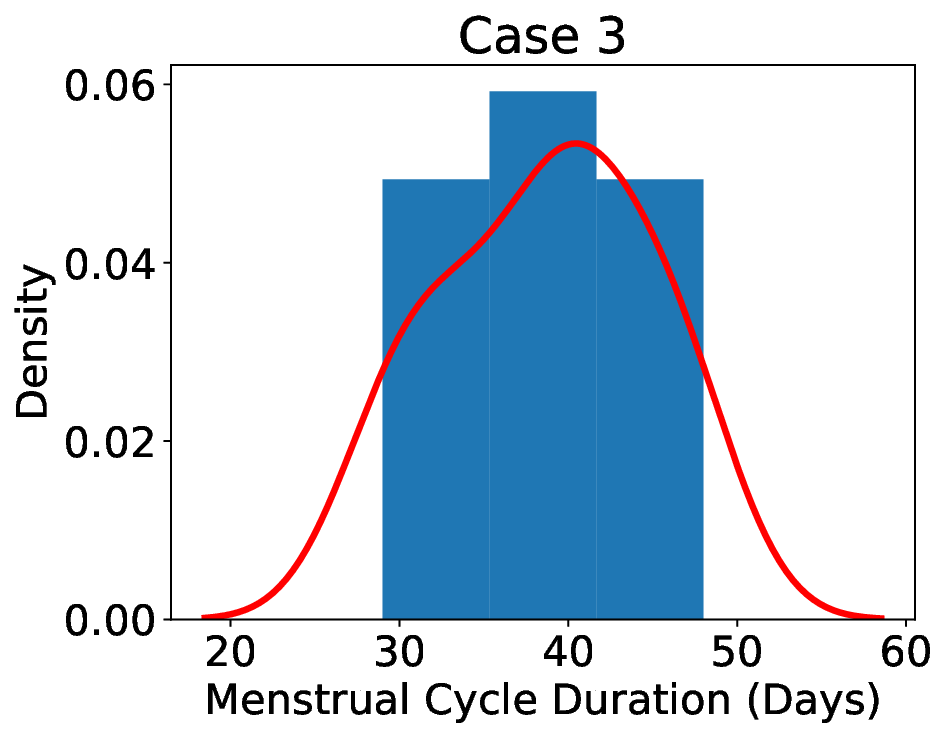}}
    \caption{Menstrual cycle distribution. }
    \label{fig1}
\end{figure}

\begin{figure}[!htb]
    \centering
 \subfigure{ \includegraphics[scale=0.3]{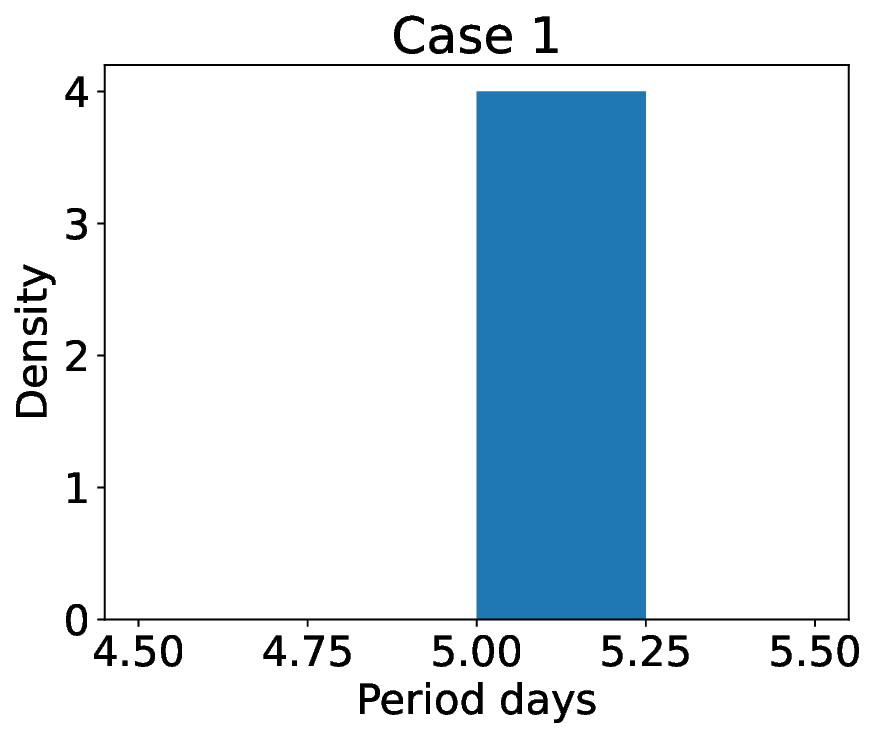}}
 \subfigure{ \includegraphics[scale=0.3]{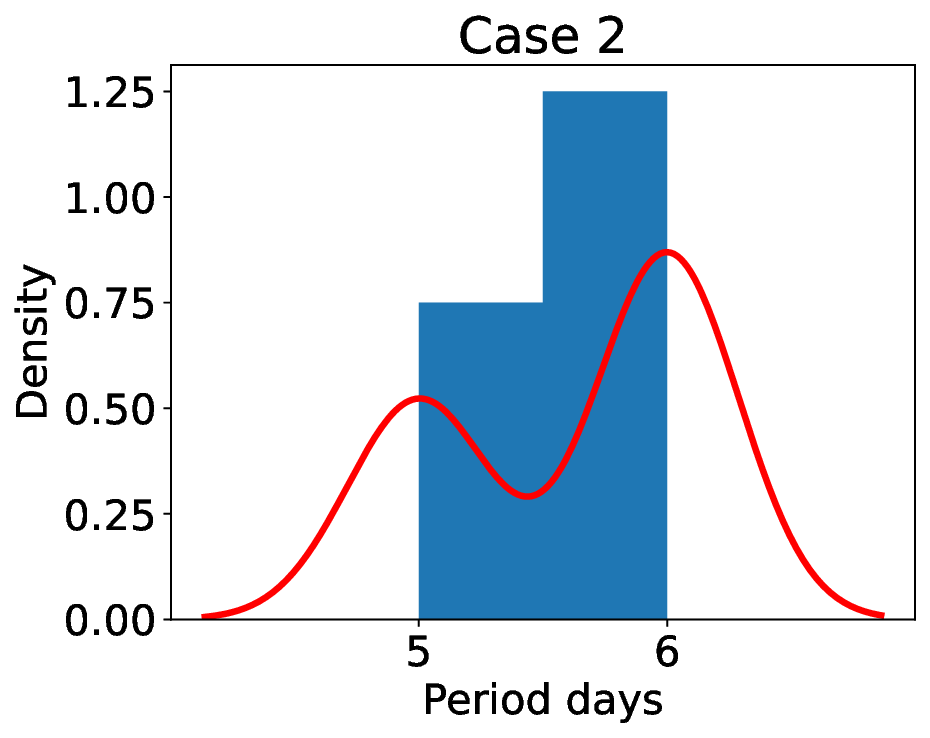}}
 \subfigure{ \includegraphics[scale=0.3]{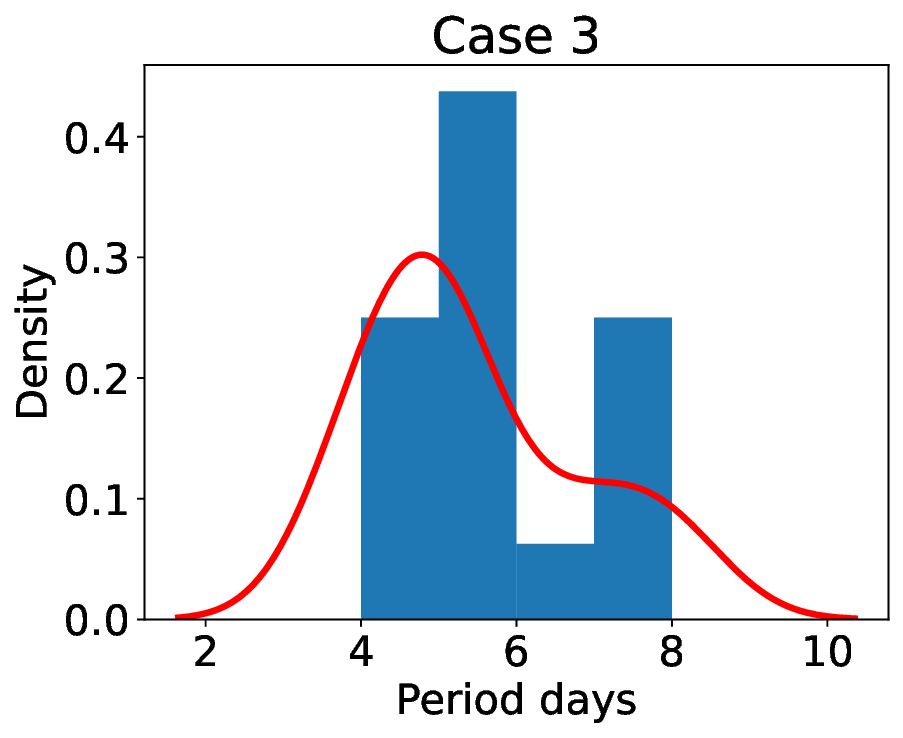}}
    \caption{Periods length distribution.}
    \label{fig3}
\end{figure}

Figures \ref{fig1} and  \ref{fig3}  show the menstrual cycle and period distribution in the three cases, respectively. As shown, in Figures \ref{fig1} and \ref{fig3}, in case 1, the menstrual cycle duration is between 28 and 30 days, and the period duration is 5 days. A large density is concentrated in cycles of 30 days, and the period duration is 5 days. 
In case 2, the menstrual cycle duration is between 28 and 34 days (Figure \ref{fig1}), and the period duration is between 5 and 6 days (Figure \ref{fig3}). In case 2, we have more cycle of 28 and 34 days , and the period duration with 6 days.  In case 3, the menstrual cycle duration is between 28 and 49 days (Figure \ref{fig1}), and the period duration is between 4 and 8 days (Figure \ref{fig3}). 

\begin{figure}[!htb]
    \centering
 \subfigure{ \includegraphics[scale=0.3]{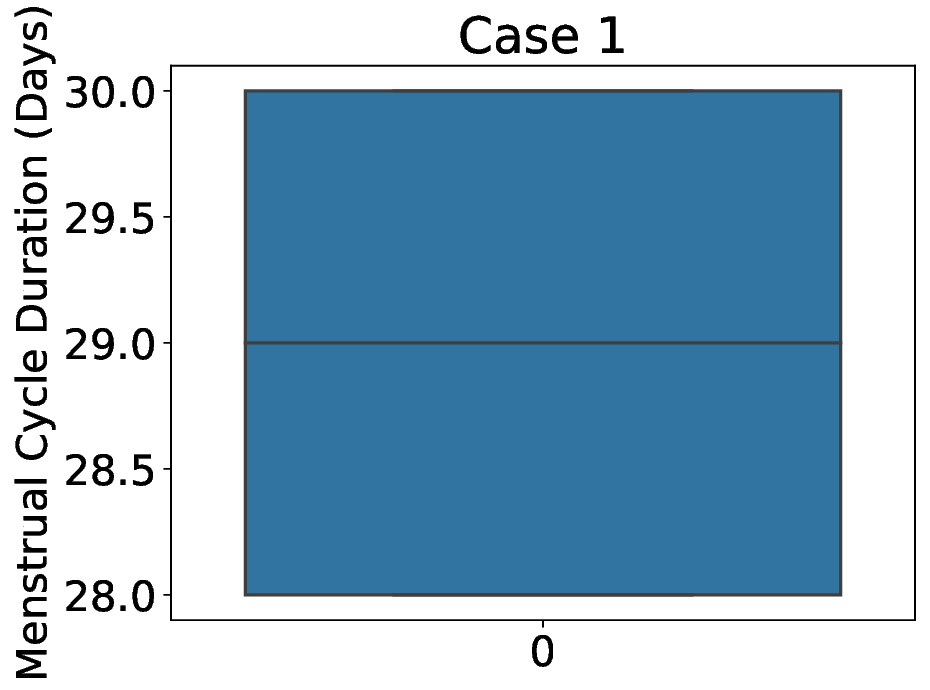}}
 \subfigure{ \includegraphics[scale=0.3]{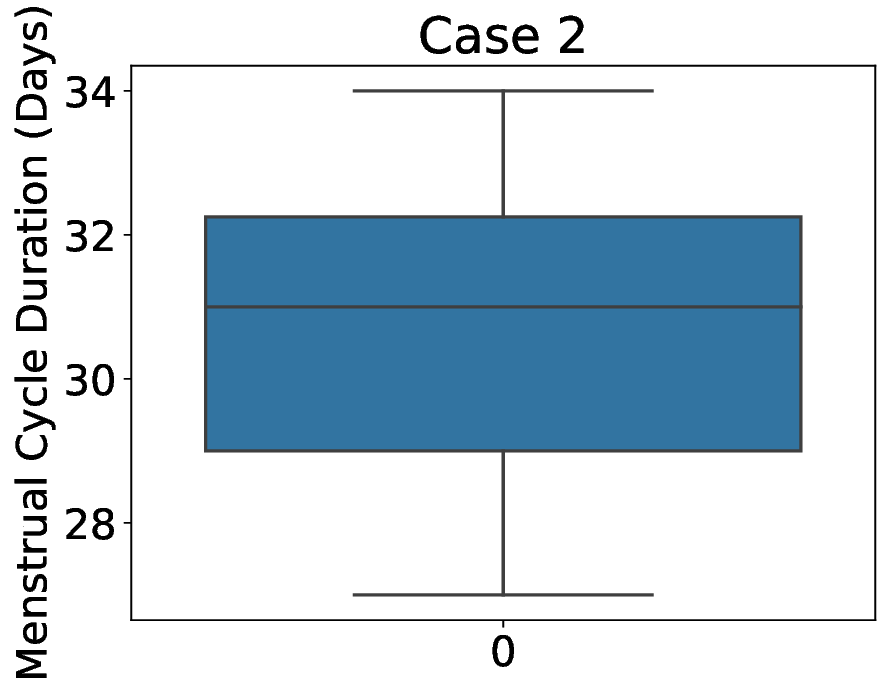}}
 \subfigure{ \includegraphics[scale=0.3]{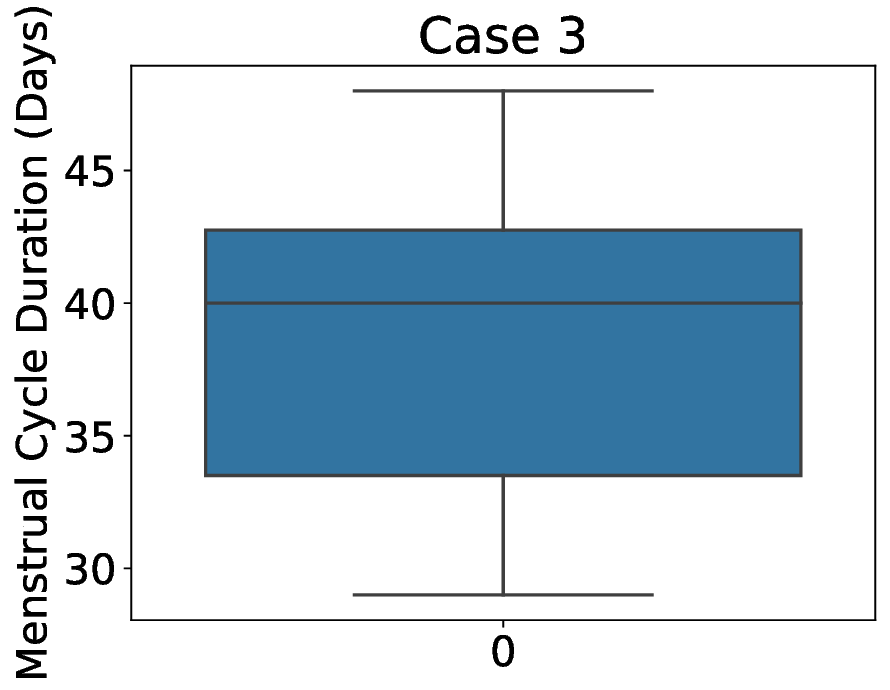}}
    \caption{Boxplot of the menstrual cycle in the three cases.}
    \label{fig2}
\end{figure}

Figure \ref{fig2} shows the boxplot of the menstrual cycle distribution in the three cases. As shown, in Figure \ref{fig2}, in case 1, the menstrual cycle duration is between 28 and 30 days, and the period duration is 5 days. In case 2, the menstrual cycle duration is between 28 and 35 days, and the period duration is between 5 and 6 days. In case 3, the menstrual cycle duration in the three cases. As shown, in Figure \ref{fig2}, in case 1, the menstrual cycle duration mean is 29 days, and the period duration mean is 5 days. In case 2, the menstrual cycle duration mean is 31 days, and the period duration mean is 5.5 days. In case 3, the menstrual cycle duration mean is 38 days, and the period duration mean is 6 days.

\begin{figure}[!htb]
    \centering
 \subfigure{ \includegraphics[scale=0.3]{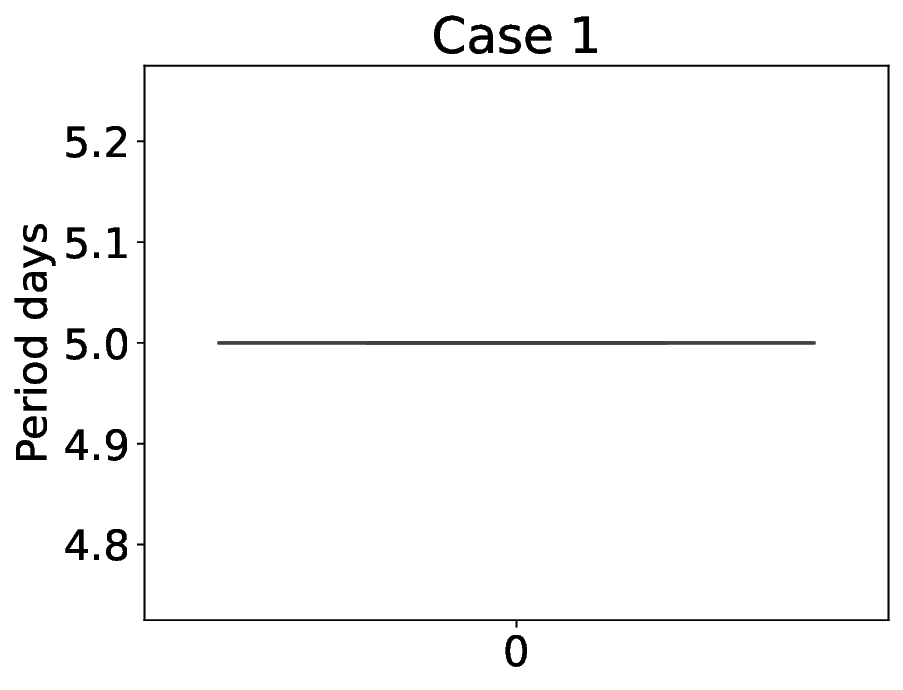}}
 \subfigure{ \includegraphics[scale=0.3]{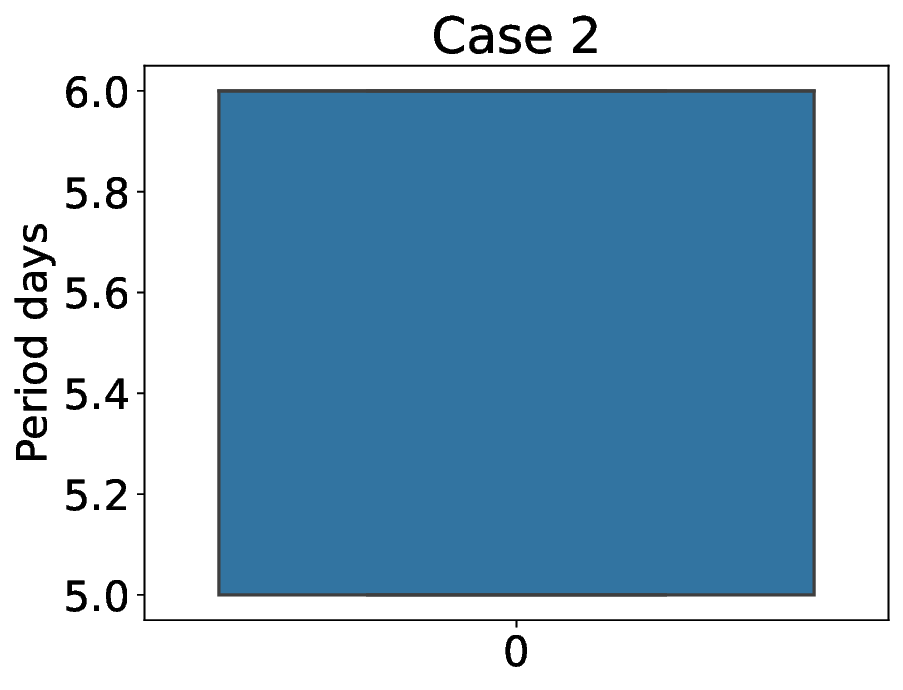}}
 \subfigure{ \includegraphics[scale=0.3]{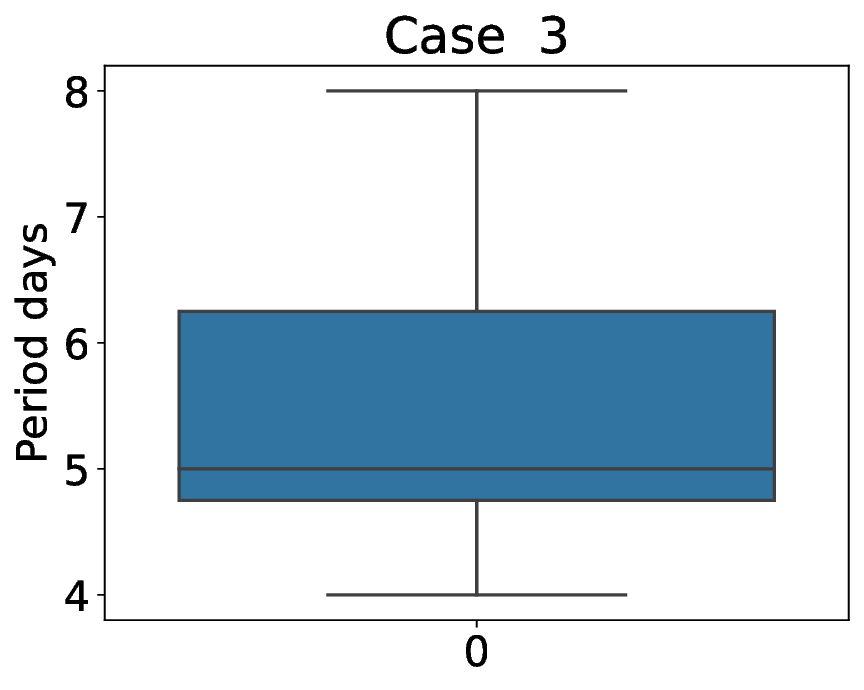}}
    \caption{Boxplot of the periods length distribution in the three cases.}
    \label{fig4}
\end{figure}

In Figures \ref{fig4} (a), (b), and (c) are depicted the boxplot of the periods length. As shown, in Figure \ref{fig4} (c), in case 3, the period length varies significantly from cycle to cycle. This can made it difficult to predict when the next period will start and how long it will last.   This variability in period length is what characterizes an irregular menstrual cycle.

With the data generated, we can now proceed with the simulation. In the following section, we will describe the simulations conducted and present the results obtained.

\section{Simulation and Results}

To perform the models' implementation, we used the TensorFlow framework version 2.12.0 for the LSTM model. We also utilized Darts library version 0.8.1 for ARIMA model implementation. Moreover, to execute Huber regression, LASSO, and OMP, we applied the PyCaret library regression module version 3.0.

We divided this section into three subsections to present the simulations conducted for three different cases: case 1 and case 2 with datasets representing regular cycles, and case 3 with datasets representing irregular cycles. 
For case 1, we implemented an LSTM network with 64 units and with hyperbolic tangent activation function. This layer processes the input sequence and generates a hidden state that summarizes the information in the sequence. Moreover, we added a dropout layer with a rate of 0.05 to prevent overfitting. In the output layer a rectified linear unit activation function was used. 

For cases 2 and 3, we utilized a stack of three LSTM layers with progressively decreasing numbers of units: 128, 64, and 32, respectively. These layers were interspersed with dropout layers set at a regularization rate of 0.2. The ultimate dense layer generates predictions using the rectified linear unit activation function.

The reliability of the models is evaluated using measures, such as mean absolute error (MAE), mean squared error (MSE),  and root mean squared error (RMSE).

\subsection{Case 1}

Table \ref{tab1} shows the models and their respective metrics, such as MAE, MSE, and RMSE,  which were calculated based on predictions made for the next 14 cycles by each model.   These metrics provide a comprehensive evaluation of the trained models. The lower the MAE, the more acceptable is the model's performance.  Because it provides a good indication of how close the predictions are to the true values. Alike to MAE, a lower MSE value indicates better model performance. Also,  like MSE, a lower RMSE signifies more suitable model performance. Out of all the models, LSTM stands out as having the smaller values in all the metrics, indicating its superior effectiveness to predict the regular menstrual cycle.

\begin{table}[!htb]
    \centering
    \begin{tabular}{lccc}
     Model   &  MAE  &   MSE  &  RMSE\\
     \hline
      LSTM   & 0.3000 &  0.5477 &   0.7401   \\
   ARIMA & 0.7400 & 0.8400 & 0.9165  \\ 

Lasso Regression & 0.7580 & 0.7649 & 0.8641   \\
Lasso Least Angle Regression & 0.7580 & 0.7649 & 0.8641   \\
Elastic Net & 0.7580 & 0.7649&  0.8641 \\
Dummy Regressor & 0.7580 & 0.7649 & 0.8641   \\
Light Gradient Boosting Machine&  0.7580 & 0.7649 & 0.8641   \\
Bayesian Ridge&  0.7583 & 0.7654  &0.8644  \\ 
Ridge Regression & 0.7882 & 0.8076 & 0.8885  \\
Linear Regression & 0.7892  &0.8092&  0.8894\\   
Least Angle Regression&  0.7892&  0.8092 & 0.8894   \\ 
Orthogonal Matching Pursuit & 0.7892&  0.8092&  0.8894 \\   
Huber Regressor & 0.7943 & 0.8174&  0.8935   \\
K Neighbors Regressor&  0.8367 & 0.9320 & 0.9374 \\  
Random Forest Regressor&  0.9610&  1.3042&  1.1017  \\ 
Passive Aggressive Regressor  &0.9692  &1.3238  &1.1040   \\
Extra Trees Regressor&  1.0076  &1.3885&  1.1392\\   
AdaBoost Regressor & 0.9652&  1.3531 & 1.1162   \\
CatBoost Regressor & 1.0093&  1.5354 & 1.1797\\   
Gradient Boosting Regressor&  1.0412  &1.6650 & 1.2364   \\
Decision Tree Regressor & 1.0750 & 1.8417 & 1.2962   \\ 
Extreme Gradient Boosting  &1.1826& 2.0305 & 1.3720\\
    \end{tabular}
    \caption{Case 1 models metrics results for predictions of next 14 cycles.}
    \label{tab1}
\end{table}

\begin{figure}[!htb]
    \centering
      \subfigure[]{\includegraphics[scale=0.5]{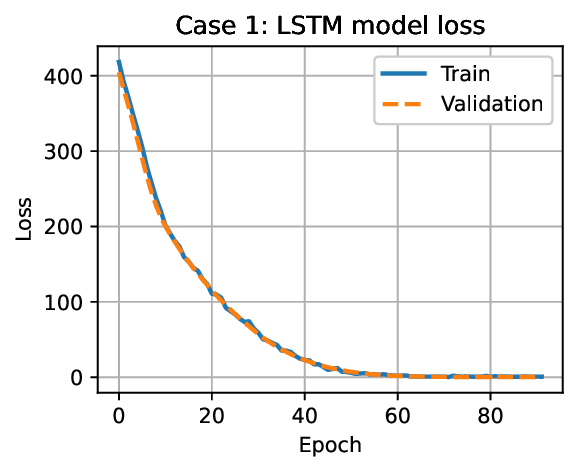}  } 
    \subfigure[]{  \includegraphics[scale=0.5]{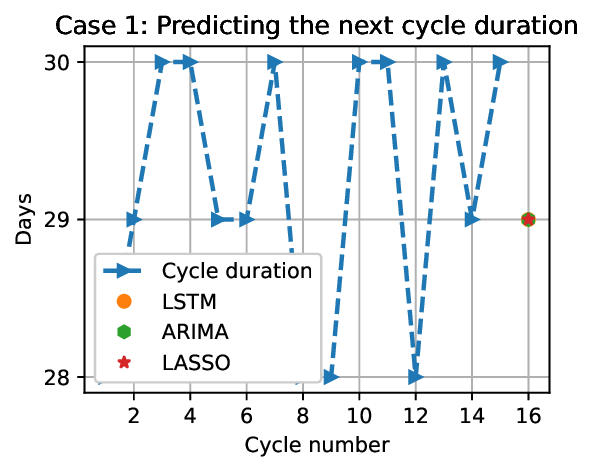}  }
     \subfigure[]{\includegraphics[scale=0.5]{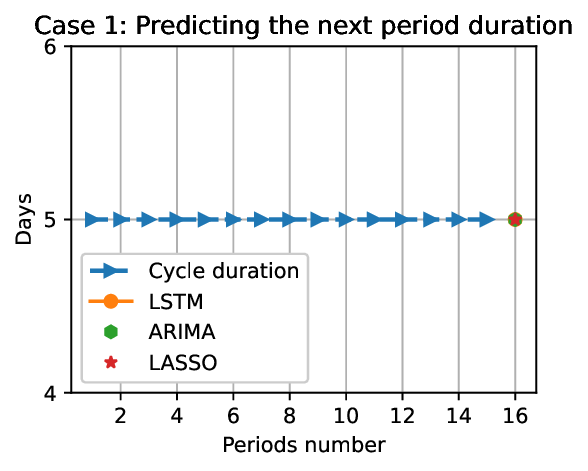}   }
    \caption{(a) Evolution of the loss function across epochs for LSTM model. (b) Prediction for time+1 of cycle time series over days. (c) Period time series over days.}
    \label{fig11}
\end{figure}

Figure \ref{fig11} (a) depict the change in the loss function value during the training process through different epochs for the LSTM network. We needed approximately 100 epochs to the model convergence.  In Figures \ref{fig11} (b) and (c),  the time-series of the cycles and periods are shown, respectively, and the prediction provided by LSTM, ARIMA, and LASSO models. Each model provided an identical prediction for the next cycle, i.e., time+1.  Moreover, we observed that there is a periodic pattern over time. 

\break
\subsection{Case 2}

Table \ref{tab2} depicts the models metrics evaluation for regular cycles. In contrast to Case 1, the Case 2 dataset exhibits greater variation in cycles. Analyzing the table, different models achieve better results compared with the results in case 1. In case 2, we had models such as Huber Regressor  and K Neighbors Regressor.  However, the LSTM model continues to exhibit the best performance across all metrics for predictions made for the next 14 cycles. 
Ridge Regression, Linear Regression, and Least Angle Regression models have the same values for MAE, MSE, and RMSE. They perform slightly worse than the Huber Regressor but still exhibit good performance.

\begin{table}[!htb]
    \centering
    \begin{tabular}{lccc}
       Model                           & MAE    & MSE & RMSE             \\
     \hline
        LSTM  & 2.1000  &  2.5495 & 1.5967 \\
Huber Regressor                 & 1.7712 & 5.0785       & 1.9973          \\
K Neighbors Regressor           & 1.8600 & 5.5167       & 2.0995           \\
Ridge Regression                & 1.8115 & 5.1098       & 2.0305         \\
Linear Regression               & 1.8119 & 5.1223       & 2.0275         \\
Least Angle Regression          & 1.8119 & 5.1223       & 2.0275          \\
Bayesian Ridge                  & 1.9036 & 5.2605       & 2.0980         \\
           ARIMA  & 2.4000  &  2.7100 & 1.6462  \\
Elastic Net                     & 2.1392 & 5.9010       & 2.2964         \\
Dummy Regressor                 & 2.1865 & 5.9786       & 2.3316          \\
Light Gradient Boosting Machine & 2.1865 & 5.9786       & 2.3316          \\
Lasso Least Angle Regression    & 2.1887 & 5.9879       & 2.3337         \\
Lasso Regression                & 2.1887 & 5.9879       & 2.3337        \\
Orthogonal Matching Pursuit     & 2.1008 & 6.1891       & 2.3154        \\
Random Forest Regressor         & 2.2672 & 7.3894       & 2.5504         \\
Gradient Boosting Regressor     & 2.7488 & 10.2011      & 3.0975          \\
CatBoost Regressor              & 2.5281 & 8.9969       & 2.8520         \\
Extra Trees Regressor           & 2.3502 & 8.5454       & 2.7297          \\
Extreme Gradient Boosting       & 2.4686 & 9.0248       & 2.8734         \\
AdaBoost Regressor              & 2.7140 & 10.9011      & 3.0415          \\
Passive Aggressive Regressor    & 2.3072 & 7.9378       & 2.5390          \\
Decision Tree Regressor         & 2.8667 & 11.5167      & 3.2344    
    \end{tabular}
    \caption{Case 2 models metrics results of predictions for the next 14 cycles.}
    \label{tab2}
\end{table}

\begin{figure}[!htb]
    \centering
      \subfigure[]{  \includegraphics[scale=0.5]{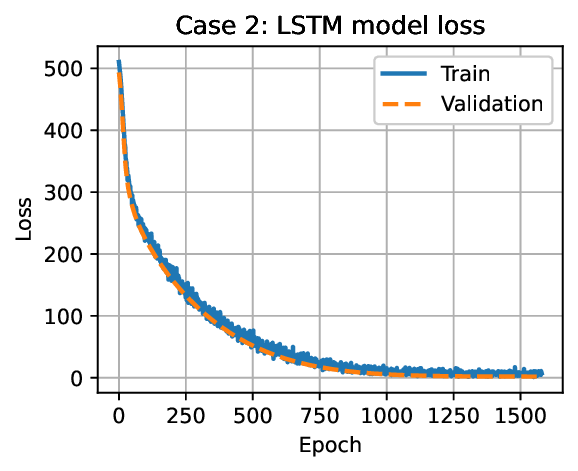}  } 
  \subfigure[]{  \includegraphics[scale=0.5]{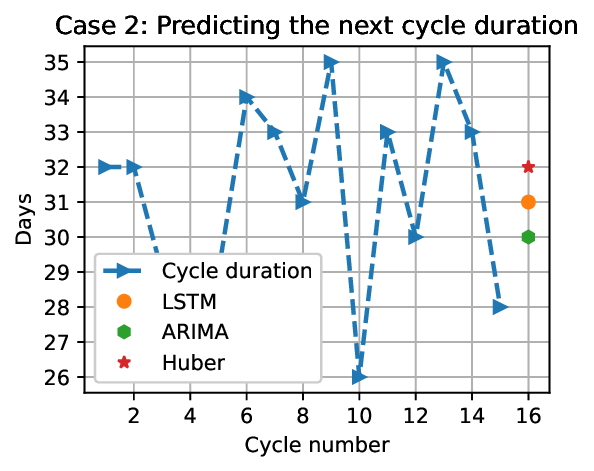}  } 
    \subfigure[]{  \includegraphics[scale=0.5]{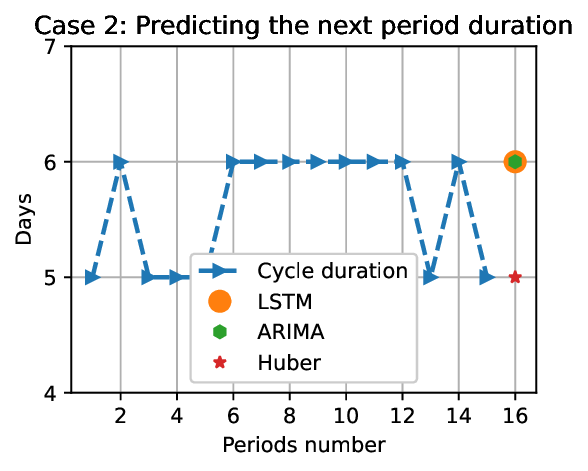}  }
    \caption{(a) Progression of the loss function throughout epochs. (b) Time series representing the cycle as a function of cycle number. (c) Time series illustrating the period as a function of period number.}
    \label{fig22}
\end{figure}

Figures \ref{fig22} (a), (b), and (c) present the simulation results for regular cycle data. The progression of the loss function over epochs during LSTM training is depicted in Figure \ref{fig22} (a). The LSTM model took around 1600 epochs to converge. Figures \ref{fig22} (b) and (c) display the time series of cycles and periods for regular cycle data, respectively. 
Differently from Case 1, in Case 2 each model provided a distinct prediction for the next cycle, i.e., cycle $16^{th}$, as shown in Figure \ref{fig22} (b). LSTM model provided a cycle of 31 days, the ARIMA model predicted a cycle of 30 days, and differently, Huber regression predicted a cycle of 32 days. The period duration provided by LSTM and ARIMA are equal, as depicted in Figure \ref{fig22}  (c). 

\break
\subsection{Case 3}

Table \ref{tab3} shows the models metrics evaluation for time series of irregular cycles. In contrast to Case 1 and Case 2, the dataset in Case 3 presents a greater variation in cycles with no perceptible patterns. Based on Table \ref{tab3}, the LSTM model performed better in therms of MAE, MSE, and RMSE. Both ARIMA and OMP models presented good performance, however, the ARIMA model is considered better than the OMP for the irregular cycle time series.

\begin{table}[!htb]
    \centering
    \begin{tabular}{lccc}
    Model                           & MAE    & MSE     & RMSE          \\
     \hline
  LSTM   &  3.4000  &  4.2895 &   2.0711   \\
  ARIMA & 7.3000 &  7.7964 & 2.7922  \\
Orthogonal Matching Pursuit     & 5.3373 & 41.1000 & 6.1243     \\
Elastic Net                     & 5.3686 & 41.1295 & 6.1588    \\
Huber Regressor                 & 5.5458 & 43.5776 & 6.2851    \\
Ridge Regression                & 5.4826 & 42.5662 & 6.2523     \\
Linear Regression               & 5.4914 & 42.7152 & 6.2604    \\
Least Angle Regression          & 5.4914 & 42.7152 & 6.2604      \\
Lasso Regression                & 5.4554 & 42.2978 & 6.2402     \\
Lasso Least Angle Regression    & 5.4554 & 42.2978 & 6.2402     \\
Dummy Regressor                 & 5.4702 & 41.5427 & 6.2063     \\
Light Gradient Boosting Machine & 5.4702 & 41.5427 & 6.2063     \\
Bayesian Ridge                  & 5.5910 & 44.0523 & 6.3834    \\
K Neighbors Regressor           & 5.6383 & 45.9650 & 6.5151     \\
Passive Aggressive Regressor    & 5.6844 & 48.3409 & 6.5683     \\
Random Forest Regressor         & 6.2947 & 59.3721 & 7.3739    \\
AdaBoost Regressor              & 6.2856 & 56.1740 & 7.3034     \\
CatBoost Regressor              & 7.2368 & 76.8913 & 8.4820    \\
Gradient Boosting Regressor     & 7.2034 & 75.0468 & 8.3428     \\
Decision Tree Regressor         & 7.1083 & 75.7250 & 8.3555    \\
Extra Trees Regressor           & 7.1378 & 69.3403 & 8.1291     \\
Extreme Gradient Boosting       & 7.5025 & 77.0432 & 8.5549    
    \end{tabular}
    \caption{Case 3 models metrics results of predictions for   the next 14 cycles.}
    \label{tab3}
\end{table}

\begin{figure}[!htb]
    \centering
      \subfigure[]{  \includegraphics[scale=0.5]{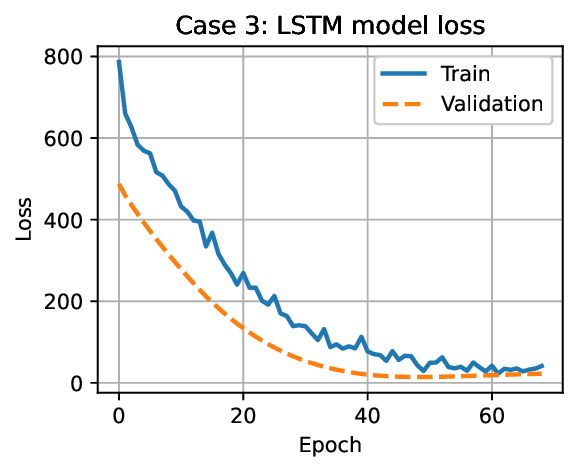}  } 
    \subfigure[]{  \includegraphics[scale=0.5]{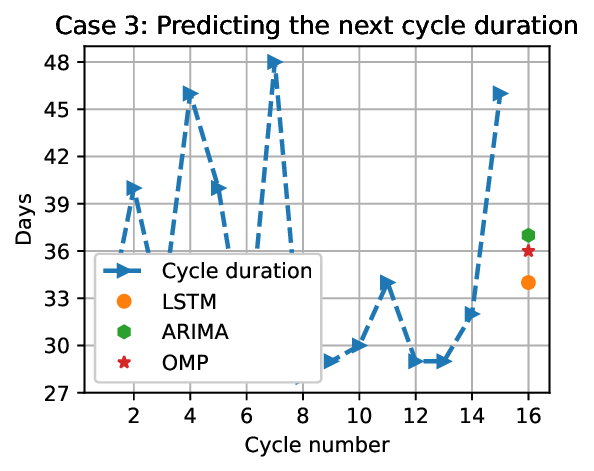}  }
     \subfigure[]{  \includegraphics[scale=0.5]{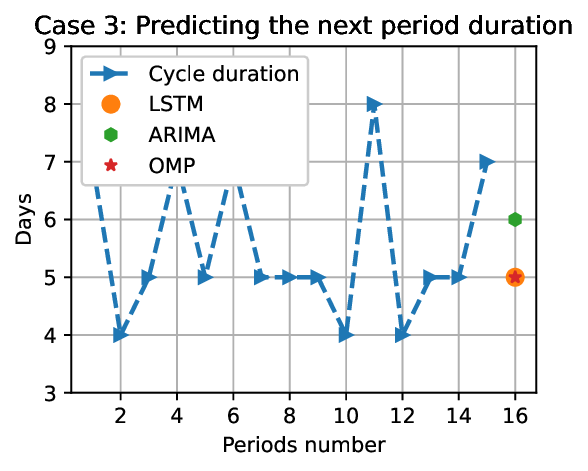}  } 
    \caption{(a) Evolution of the loss function over epochs. (b) Cycle time series as a function of cycle number. (c) Period time series as a function of period number.}
    \label{fig33}
\end{figure}

Figures \ref{fig33} (a), (b), and (c) show the simulation results for irregular cycle data. Figure \ref{fig33} (a) displays the progression of the loss function over epochs during LSTM training. Figure \ref{fig33} (b) show the cycle time series and the predictions provided by models, and Figure \ref{fig33}  (c) depicts the period time series as a function of period number and the forecasts given by models.

Based on the given results, the LSTM network model appears to be the best in terms of error metrics. Also, the ARIMA model has relatively low values for MAE, MSE, and RMSE. While not as good as the LSTM model, the ARIMA model performs better than most other models, this is due to the flexibility of the ARIMA model, it can capture both short-term and long-term dependencies in the data. However, the LSTM model has the capability to capture temporal relationships within cycle time series.

\break
\section{Conclusion}

In this study, we applied machine learning algorithms to predict menstrual cycle using data from generated by defined model. The algorithms can  identify patterns and correlations that may not be immediately apparent to humans. This can lead to new insights and discoveries in the field of menstrual cycle prediction. Moreover, our results suggest that machine learning models can accurately predict menstrual cycle phase with small error. Our findings have important implications for women's health and may be used to inform personalized reproductive health decisions, such as family planning and fertility treatment. In addition, the models can be trained on individualized data, allowing for personalized predictions of menstrual cycle patterns. This could be particularly useful for women with irregular cycles, as traditional prediction methods may not work as well for them.
Future work in this area could focus on refining our models by incorporating additional data sources, such as  menstrual symptoms, hormonal data, and conducting larger-scale studies to validate our findings.

\bibliographystyle{abbrv}
\bibliography{main}

\begin{thebibliography}{10}

\bibitem{alhussein2018voice}
M.~Alhussein and G.~Muhammad.
\newblock Voice pathology detection using deep learning on mobile healthcare
  framework.
\newblock {\em IEEE Access}, 6:41034--41041, 2018.

\bibitem{bull2019real}
J.~R. Bull, S.~P. Rowland, E.~B. Scherwitzl, R.~Scherwitzl, K.~G. Danielsson,
  and J.~Harper.
\newblock Real-world menstrual cycle characteristics of more than 600,000
  menstrual cycles.
\newblock {\em NPJ digital medicine}, 2(1):1--8, 2019.

\bibitem{cai2011orthogonal}
T.~T. Cai and L.~Wang.
\newblock Orthogonal matching pursuit for sparse signal recovery with noise.
\newblock {\em IEEE Transactions on Information theory}, 57(7):4680--4688,
  2011.

\bibitem{campbell2021menstrual}
L.~R. Campbell, A.~L. Scalise, B.~T. DiBenedictis, and S.~Mahalingaiah.
\newblock Menstrual cycle length and modern living: a review.
\newblock {\em Current opinion in endocrinology, diabetes, and obesity},
  28(6):566, 2021.

\bibitem{chatfield2000time}
C.~Chatfield.
\newblock {\em Time-series forecasting}.
\newblock CRC press, 2000.

\bibitem{chen2014mathematical}
C.~Chen and J.~P. Ward.
\newblock A mathematical model for the human menstrual cycle.
\newblock {\em Mathematical medicine and biology: a journal of the IMA},
  31(1):65--86, 2014.

\bibitem{de2021modelling}
T.~de~Paula~Oliveira, G.~Bruinvels, C.~R. Pedlar, B.~Moore, and J.~Newell.
\newblock Modelling menstrual cycle length in athletes using state-space
  models.
\newblock {\em Scientific reports}, 11(1):1--14, 2021.

\bibitem{denny2019hope}
A.~Denny, A.~Raj, A.~Ashok, C.~M. Ram, and R.~George.
\newblock i-hope: Detection and prediction system for polycystic ovary syndrome
  (pcos) using machine learning techniques.
\newblock In {\em TENCON 2019-2019 IEEE Region 10 Conference (TENCON)}, pages
  673--678. IEEE, 2019.

\bibitem{deverashetti2022security}
M.~Deverashetti, K.~Ranjitha, and K.~Pradeepthi.
\newblock Security analysis of menstruation cycle tracking applications using
  static, dynamic and machine learning techniques.
\newblock {\em Journal of Information Security and Applications}, 67:103171,
  2022.

\bibitem{fukaya2017forecasting}
K.~Fukaya, A.~Kawamori, Y.~Osada, M.~Kitazawa, and M.~Ishiguro.
\newblock The forecasting of menstruation based on a state-space modeling of
  basal body temperature time series.
\newblock {\em Statistics in Medicine}, 36(21):3361--3379, 2017.

\bibitem{harlow1995epidemiology}
S.~D. Harlow and S.~A. Ephross.
\newblock Epidemiology of menstruation and its relevance to women's health.
\newblock {\em Epidemiologic reviews}, 17(2):265--286, 1995.

\bibitem{hochreiter1997long}
S.~Hochreiter and J.~Schmidhuber.
\newblock Long short-term memory.
\newblock {\em Neural computation}, 9(8):1735--1780, 1997.

\bibitem{huber1992robust}
P.~J. Huber.
\newblock Robust estimation of a location parameter.
\newblock {\em Breakthroughs in statistics: Methodology and distribution},
  pages 492--518, 1992.

\bibitem{kallummil2018signal}
S.~Kallummil and S.~Kalyani.
\newblock Signal and noise statistics oblivious orthogonal matching pursuit.
\newblock In {\em International Conference on Machine Learning}, pages
  2429--2438. PMLR, 2018.

\bibitem{kleinschmidt2019advantages}
T.~K. Kleinschmidt, J.~R. Bull, V.~Lavorini, S.~P. Rowland, J.~T. Pearson,
  E.~B. Scherwitzl, R.~Scherwitzl, and K.~G. Danielsson.
\newblock Advantages of determining the fertile window with the individualised
  natural cycles algorithm over calendar-based methods.
\newblock {\em The European Journal of Contraception \& Reproductive Health
  Care}, 24(6):457--463, 2019.

\bibitem{kwak2018irregular}
Y.~Kwak and Y.~Kim.
\newblock Irregular menstruation according to occupational status.
\newblock {\em Women \& Health}, 58(10):1135--1150, 2018.

\bibitem{li2021generative}
K.~Li, I.~Urteaga, A.~Shea, V.~J. Vitzthum, C.~H. Wiggins, and N.~Elhadad.
\newblock A generative, predictive model for menstrual cycle lengths that
  accounts for potential self-tracking artifacts in mobile health data.
\newblock {\em arXiv preprint arXiv:2102.12439}, 2021.

\bibitem{li2022predictive}
K.~Li, I.~Urteaga, A.~Shea, V.~J. Vitzthum, C.~H. Wiggins, and N.~Elhadad.
\newblock A predictive model for next cycle start date that accounts for
  adherence in menstrual self-tracking.
\newblock {\em Journal of the American Medical Informatics Association},
  29(1):3--11, 2022.

\bibitem{munro2012figo}
M.~G. Munro, H.~O. Critchley, and I.~S. Fraser.
\newblock The figo systems for nomenclature and classification of causes of
  abnormal uterine bleeding in the reproductive years: who needs them?
\newblock {\em American journal of obstetrics and gynecology}, 207(4):259--265,
  2012.

\bibitem{nguyen2021detecting}
B.~T. Nguyen, R.~D. Pang, A.~L. Nelson, J.~T. Pearson, E.~Benhar~Noccioli,
  H.~R. Reissner, A.~Kraker~von Schwarzenfeld, and J.~Acuna.
\newblock Detecting variations in ovulation and menstruation during the
  covid-19 pandemic, using real-world mobile app data.
\newblock {\em PloS one}, 16(10):e0258314, 2021.

\bibitem{nithya2017predictive}
B.~Nithya and V.~Ilango.
\newblock Predictive analytics in health care using machine learning tools and
  techniques.
\newblock In {\em 2017 International Conference on Intelligent Computing and
  Control Systems (ICICCS)}, pages 492--499. IEEE, 2017.

\bibitem{pearson2021natural}
J.~T. Pearson, M.~Chelstowska, S.~P. Rowland, E.~Mcilwaine, E.~Benhar,
  E.~Berglund~Scherwitzl, S.~Walker, K.~Gemzell~Danielsson, and R.~Scherwitzl.
\newblock Natural cycles app: contraceptive outcomes and demographic analysis
  of uk users.
\newblock {\em The European Journal of Contraception \& Reproductive Health
  Care}, 26(2):105--110, 2021.

\bibitem{pierson2018modeling}
E.~Pierson, T.~Althoff, and J.~Leskovec.
\newblock Modeling individual cyclic variation in human behavior.
\newblock In {\em Proceedings of the 2018 World Wide Web Conference}, pages
  107--116, 2018.

\bibitem{rahane2018lung}
W.~Rahane, H.~Dalvi, Y.~Magar, A.~Kalane, and S.~Jondhale.
\newblock Lung cancer detection using image processing and machine learning
  healthcare.
\newblock In {\em 2018 International Conference on Current Trends towards
  Converging Technologies (ICCTCT)}, pages 1--5. IEEE, 2018.

\bibitem{rees2005abnormal}
M.~Rees, S.~L. Hope, and V.~A. Ravnikar.
\newblock {\em The Abnormal Menstrual Cycle}.
\newblock CRC Press, 2005.

\bibitem{roblitz2013mathematical}
S.~R{\"o}blitz, C.~St{\"o}tzel, P.~Deuflhard, H.~M. Jones, D.-O. Azulay, P.~H.
  van~der Graaf, and S.~W. Martin.
\newblock A mathematical model of the human menstrual cycle for the
  administration of gnrh analogues.
\newblock {\em Journal of theoretical biology}, 321:8--27, 2013.

\bibitem{rostvik2022cash}
C.~M. R{\o}stvik.
\newblock {\em Cash Flow: The businesses of menstruation}.
\newblock UCL Press, 2022.

\bibitem{sarwar2018prediction}
M.~A. Sarwar, N.~Kamal, W.~Hamid, and M.~A. Shah.
\newblock Prediction of diabetes using machine learning algorithms in
  healthcare.
\newblock In {\em 2018 24th international conference on automation and
  computing (ICAC)}, pages 1--6. IEEE, 2018.

\bibitem{setton2016accuracy}
R.~A. Setton, C.~H. Tierney, and T.~Tsai.
\newblock The accuracy of websites and cellular phone applications in
  predicting the fertile window [12g].
\newblock {\em Obstetrics \& Gynecology}, 127:62S, 2016.

\bibitem{shinde2018intelligent}
S.~A. Shinde and P.~R. Rajeswari.
\newblock Intelligent health risk prediction systems using machine learning: a
  review.
\newblock {\em International Journal of Engineering \& Technology},
  7(3):1019--1023, 2018.

\bibitem{tatsumi2020age}
T.~Tatsumi, M.~Sampei, K.~Saito, Y.~Honda, Y.~Okazaki, N.~Arata, K.~Narumi,
  N.~Morisaki, T.~Ishikawa, and S.~Narumi.
\newblock Age-dependent and seasonal changes in menstrual cycle length and body
  temperature based on big data.
\newblock {\em Obstetrics and Gynecology}, 136(4):666, 2020.

\bibitem{tibshirani1996regression}
R.~Tibshirani.
\newblock Regression shrinkage and selection via the lasso.
\newblock {\em Journal of the Royal Statistical Society: Series B
  (Methodological)}, 58(1):267--288, 1996.

\bibitem{tiwari2022sposds}
S.~Tiwari, L.~Kane, D.~Koundal, A.~Jain, A.~Alhudhaif, K.~Polat, A.~Zaguia,
  F.~Alenezi, and S.~A. Althubiti.
\newblock Sposds: A smart polycystic ovary syndrome diagnostic system using
  machine learning.
\newblock {\em Expert Systems with Applications}, 203:117592, 2022.

\bibitem{trickey2004women}
R.~Trickey.
\newblock {\em Women, Hormones and the Menstrual Cycle: herbal and medical
  solutions from adolescence to menopause}.
\newblock Allen \& Unwin, 2004.

\bibitem{urteaga2021generative}
I.~Urteaga, K.~Li, A.~Shea, V.~J. Vitzthum, C.~H. Wiggins, and N.~Elhadad.
\newblock A generative modeling approach to calibrated predictions: a use case
  on menstrual cycle length prediction.
\newblock In {\em Machine Learning for Healthcare Conference}, pages 535--566.
  PMLR, 2021.

\bibitem{yu2022tracking}
J.-L. Yu, Y.-F. Su, C.~Zhang, L.~Jin, X.-H. Lin, L.-T. Chen, H.-F. Huang, and
  Y.-T. Wu.
\newblock Tracking of menstrual cycles and prediction of the fertile window via
  measurements of basal body temperature and heart rate as well as
  machine-learning algorithms.
\newblock {\em Reproductive Biology and Endocrinology}, 20(1):1--12, 2022.

\end{thebibliography}
\end{document}